\documentclass{article}
\usepackage{PRIMEarxiv}
\usepackage[utf8]{inputenc} %
\usepackage[T1]{fontenc}    %
\usepackage{nameref, hyperref}       %
\usepackage{url}            %
\usepackage{booktabs}       %
\usepackage{amsfonts}       %
\usepackage{nicefrac}       %
\usepackage{microtype}      %
\usepackage{fancyhdr}       %
\usepackage{graphicx}
\graphicspath{{./media/}}     %
\usepackage{array}
\usepackage{afterpage}
\usepackage{ragged2e}
\usepackage{float}
\usepackage{breakcites}
\usepackage{biblatex}
\newcommand\papertitle{Predicting House Rental Prices in Ghana Using Machine Learning}
\newcommand\paperauthor{{Adzanoukpe, P.: }}

\title{\papertitle}
\author{
  Philip Adzanoukpe \\
  Epigos AI  \\
  Accra\\
  \texttt{philip@epigos.ai} \\
}
\date{December 2024}
\begin{document}
\maketitle
\begin{abstract}
This study investigates the efficacy of machine learning models for predicting house rental prices in Ghana, addressing the need for accurate and accessible housing market information. Utilising a comprehensive dataset of rental listings, we trained and evaluated various models, including CatBoost, XGBoost, and Random Forest. CatBoost emerged as the best-performing model, achieving an \(R^2\) of 0.876, demonstrating its ability to effectively capture complex relationships within the housing market. Feature importance analysis revealed that location-based features, number of bedrooms, bathrooms, and furnishing status are key drivers of rental prices. Our findings provide valuable insights for stakeholders, including real estate professionals, investors, and policymakers, while also highlighting opportunities for future research, such as incorporating temporal data and exploring regional variations.
\end{abstract}

\keywords{House Rental Prices \and Machine Learning \and Ghana \and Real Estate \and Regression models}

\section{Introduction}

The housing market in Ghana has been facing significant challenges, with the rental sector being particularly affected by issues such as the advance rent system, asymmetrical perceptions between landlords and tenants, and the lack of an institutional framework for regulating the market  \parencite{godwin_arku_511fecef}. These challenges create a highly dynamic and often opaque rental environment, where both tenants and landlords face difficulties in determining fair rental prices. This issue is further exacerbated by the absence of comprehensive and up-to-date data on rental trends, making it challenging for stakeholders to make informed decisions.

In recent years, the use of machine learning in real estate has gained traction globally as a means to address such challenges. Machine learning (ML) models can analyse large datasets, uncover hidden patterns, and make accurate predictions, thereby providing valuable insights for various stakeholders in the housing market. This paper aims to bridge the gap between the challenges in Ghana’s rental market and the capabilities of machine learning by developing a model that predicts house rental prices accurately.

This study is motivated by the need for reliable information on housing prices in Ghana, which can assist individuals in identifying fair rental rates and policymakers in formulating strategies to address housing issues \parencite{lewis_abedi_asante_4375e27e}. By leveraging machine learning techniques, this research seeks to offer a data-driven approach to addressing the complexities of the rental housing market in Ghana.

The scope of this paper includes an analysis of the current state of the rental housing market in Ghana, a comparison of different machine learning algorithms for predicting house rental prices, and the development of a predictive model. The outcomes of this research are intended to serve as a tool for improving transparency and efficiency in Ghana’s rental market, ultimately aiding both tenants and landlords in making informed decisions.

\section{Literature Review}

\subsection{Housing Market in Ghana}

The Ghanaian housing market, particularly the rental sector, faces unique challenges. These include the prevalent advance rent system  \parencite{lewis_abedi_asante_4375e27e}, often requiring tenants to pay significant upfront sums, creating financial strain. Furthermore, the lack of robust regulatory frameworks and accessible data contributes to market opacity and potential disputes between landlords and tenants  \parencite{lewis_abedi_asante_4375e27e}. Existing research emphasises the need for data-driven solutions and policy interventions to address these issues  \parencite{na_boamah_0ba98f87}. Studies like  \parencite{de_graft_owusu_manu_5479e7f0} explore the influence of housing attributes on prices, providing valuable insights for understanding market dynamics.  \parencite{na_boamah_0ba98f87} focuses on housing affordability in Kumasi and Tamale, highlighting regional disparities and the impact of income levels and interest rates on affordability. These studies underscore the complexity of the Ghanaian housing market and the need for comprehensive data analysis to inform effective policies.
\subsection{Machine Learning for House Price Prediction}

Machine learning has emerged as a powerful tool for analysing complex datasets and making accurate predictions in various domains, including real estate. Globally, studies have demonstrated the effectiveness of machine learning algorithms in predicting house prices  \parencite{t__dhar_ba98ddee}. Algorithms such as XGBoost, CatBoost, Random Forest, Linear Regression, and Support Vector Regressor have been successfully applied to predict house prices with high accuracy  \parencite{maida_ahtesham_696fb14f} \parencite{t__dhar_ba98ddee}. These models leverage various features, including location, size, amenities, and market trends, to estimate property values. While  \parencite{house_price_prediction_using_machine_learning_1eb25409} mentions a technical issue unrelated to house price prediction, the other sources demonstrate the potential of machine learning to address the challenges of transparency and efficiency in housing markets.
\subsection{Bridging the Gap}

This research aims to bridge the gap between the specific challenges of the Ghanaian housing market and the capabilities of machine learning. By applying these techniques to data collected from Tonaton.com, this study seeks to develop a predictive model for house rental prices in Ghana. This model can contribute to increased transparency and empower both tenants and landlords with data-driven insights for informed decision-making. This research builds upon the existing literature on the Ghanaian housing market and leverages the advancements in machine learning for house price prediction to address the need for reliable and accessible housing information in Ghana.
\section{Methodology}

To conduct the research, we used housing data from Tonaton.com, a marketplace website. The data was cleaned and transformed, and various machine learning techniques such as XGBoost, CatBoost, Random Forest, Linear Regression, and Support Vector Regressor were compared. The final dataset consists of 17,890 records (houses) with 37 variables.

Additionally, we discussed the regression techniques used for predicting house prices in the subsections below, focusing on the selection of appropriate ML algorithms as supported by existing literature.
\subsection{Linear Regression}

Linear regression  \parencite{linear_regression_b8d8b5a0} is a fundamental statistical method used to model the relationship between a dependent variable and one or more independent variables by fitting a linear equation to observed data. In simple linear regression, with one independent variable, the model assumes a straight-line relationship:
\begin{equation}
y = \beta_0 + \beta_1 x + \epsilon
\end{equation}
Where:
\begin{itemize}
\item $y$ is the dependent variable.
\item $x$ is the independent variable.
\item $\beta_0$ is the y-intercept, representing the value of $y$ when x is zero.
\item $\beta_1$ is the slope, representing the change in $y$ for a one-unit change in $x$.
\item $\epsilon$ is the error term, representing the difference between the observed value of $y$ and the predicted value by the model.
\end{itemize}

\subsection{Random Forest}

Random Forest Regression  \parencite{matthias_schonlau_d5919a07} is an ensemble learning method that builds a multitude of decision trees during training and outputs the mean/average prediction of the individual trees for regression tasks. It operates by constructing multiple decision trees on bootstrapped samples of the data and using random subsets of features for each tree. This process reduces over-fitting and improves the model's ability to generalise to unseen data.  \parencite{l__nelson_sanchez_pinto_3cd7ab1e} mentions random forests in the context of feature selection, highlighting their utility in identifying important predictors.

Unlike linear regression, there isn't a single equation that represents a random forest model. The prediction for a given input x is obtained by averaging the predictions of all individual trees:
\begin{equation}
\hat{y}(x) = \frac{1}{B} \sum_{b=1}^{B} T_b(x)
\end{equation}
Where:
\begin{itemize}
\item $\hat{y}(x)$ is the predicted value for input $x$.
\item $B$ is the number of trees in the forest.
\item $Tb(x)$ is the prediction of the $b-th$ tree for input $x$.
\end{itemize}

Each tree's prediction is based on a series of decisions based on feature splits, making the overall model complex and non-linear.  \parencite{rapha_l_couronn__82e5e882} compares random forests to logistic regression, demonstrating the potential advantages of random forests in prediction accuracy, particularly in complex datasets.

\subsection{Support Vector Machines}

Support Vector Regression (SVR)  \parencite{s_r__gunn_1d50f753} aims to find the best-fitting function within a defined margin of tolerance around the data points, rather than directly fitting a line like linear regression. It utilises kernel functions to map the data into a higher-dimensional feature space where linear separation is possible. 

The equation for SVR can be expressed as:
\begin{equation}
f(x) = w^T \phi(x) + b
\end{equation}
Where:
\begin{itemize}
\item $f(x)$ is the predicted value for input $x$.
\item $w$ is the weight vector.
\item $\phi(x)$ is the feature mapping function, often non-linear.
\item $b$ is the bias term.
\end{itemize}

The goal of SVR is to minimise the error within a defined margin epsilon, while also minimising the complexity of the model, controlled by a regularisation parameter C.  \parencite{vladimir_vapnik_85193362} delves into the theoretical foundations of SVR, including the use of kernel functions and the optimisation problem involved in finding the optimal hyperplane. 

\subsection{Extreme Gradient Boosting (XGBoost)}

XGBoost  \parencite{tianqi_chen_43bb99df} is a powerful gradient boosting algorithm known for its speed and performance. It's a decision-tree-based ensemble method that works by iteratively building trees, each correcting the errors of its predecessors.  \parencite{s__premanand_81624419} briefly mentions XGBoost as a boosting algorithm derived from decision trees, highlighting its improved accuracy compared to traditional algorithms.

XGBoost combines a set of weak learners (typically decision trees) to create a strong learner. The prediction is made by adding the predictions of individual trees, weighted by their importance:
\begin{equation}
\hat{y}_i=\sum{k=1}^K f_k(x_i)
\end{equation}
Where:
\begin{itemize}
\item $\hat{y}_i$ is the prediction for the $i-th$ instance.
\item $K$ is the number of trees.
\item $f_k(x_i)$ is the prediction of the $k-th$ tree for input $x_i$.
\end{itemize}

The algorithm minimises a regularised objective function, which combines a loss function (measuring the difference between predictions and actual values) and a regularisation term (controlling model complexity).

\subsection{CatBoost}

CatBoost  \parencite{_0d710c8d} is a gradient boosting algorithm known for its effective handling of categorical features. It uses a novel technique called ordered boosting to reduce over-fitting and improve prediction accuracy.  \parencite{liudmila_prokhorenkova_a9fa0d41} details CatBoost's methodology, including its approach to handling categorical variables and its focus on reducing bias.  \parencite{shashi_bhushan_jha_db57e6b1} mentions CatBoost in the context of housing market prediction, highlighting its ability to handle categorical features effectively.

Similar to other boosting algorithms, CatBoost combines multiple decision trees to make a prediction. The prediction is a weighted sum of individual tree predictions:
\begin{equation}
\hat{y}_i=\sum{k=1}^K w_k \cdot T_k(x_i)
\end{equation}
Where:
\begin{itemize}
\item $\hat{y}_i$ is the prediction for the $i-th$ instance.
\item $K$ is the number of trees.
\item $w_k$ is the weight of the $k-th$ tree.
\item $T_k(x_i)$ is the prediction of the $k-th$ tree for input $x_i$.
\end{itemize}

 \parencite{abhishek_gupta_afe862a6} mentions CatBoost's superior performance compared to other boosting algorithms in a specific application related to polycystic ovary syndrome prognostication.
 
\subsection{Implementation}

\begin{enumerate}
\item \textbf{Data Collection:} We collected data from Tonaton.com using web scraping techniques to gather property rental listings.
\item \textbf{Data Cleaning:} We cleaned the data by removing rows with null values and regrouping listing categories.
\item \textbf{Feature Engineering:} Several feature engineering steps were performed:
\begin{itemize}
\item Mapping locations to geolocation coordinates using the Google Geolocation API.
\item One-hot encoding of categorical fields such as apartment type, condition, and furnishing.
\item Text embedding using count vectorisation to transform the top 20 amenities features.
\item Outlier removal to eliminate inconsistent prices outside normal ranges, guided by visualisation of the price distribution.
\end{itemize}
\item \textbf{Model Training:} The preprocessed data was split into training and validation sets at an 80/20 ratio. We used the scikit-learn (sklearn) Python package to implement the machine learning models. Each model was trained on the training set and validated on the test set.
\item \textbf{Hyperparameter Tuning:} Some models were further fine-tuned using grid search and random search hyper-parameter optimisation with k-fold cross-validation to select the optimal model.
\end{enumerate}

\subsection{Model Evaluation}

We evaluated the regression models using four key metrics: R-squared (R²), Mean Squared Error, Mean Absolute Error, and Root Mean Squared Error. R² served as the primary metric for hyper-parameter optimisation and cross-validation.

\subsubsection{R-squared (\texorpdfstring{\(R^2\)})}
This metric represents the proportion of variance in the dependent variable explained by the model. A higher R² indicates a better fit, with 1 being a perfect fit. However, R² alone doesn't reveal bias and can be misleading in cases of over-fitting. It's calculated as:
\begin{equation}
R^2 = 1 - \frac{SSR}{SST}
\end{equation}
where $SSR$ is the sum of squared residuals and $SST$ is the total sum of squares.
\subsubsection{Mean Squared Error (MSE}
MSE calculates the average of the squared differences between predicted and actual values. It penalises larger errors more heavily. A lower MSE suggests better model performance.
\begin{equation}
MSE = \frac{1}{n} \sum_{i=1}^{n} (y_i - \hat{y}_i)^2
\end{equation}
where $y_i$ are the actual values, $\hat{y}_i$ are the predicted values, and $n$ is the number of data points.
\subsubsection{Mean Absolute Error (MAE}
MAE calculates the average of the absolute differences between predicted and actual values. It's less sensitive to outliers than MSE. A lower MAE indicates better performance.
\begin{equation}
MAE = \frac{1}{n} \sum_{i=1}^{n} |y_i - \hat{y}_i|
\end{equation}
\subsubsection{Root Mean Squared Error (RMSE}
RMSE is the square root of MSE. It's often preferred over MSE as it's in the same units as the dependent variable, making interpretation easier.
\begin{equation}
RMSE = \sqrt{MSE}
\end{equation}

\subsection{Data Analysis}

To understand the factors influencing rental prices in Ghana, we analysed the preprocessed data using various visualisation techniques, including bar charts, pie charts, and box plots. These visualisations helped us explore feature correlations and relationships, especially with the target variable—rental price.

\begin{figure}[H]
    \centering
    \includegraphics[width=0.75\linewidth]{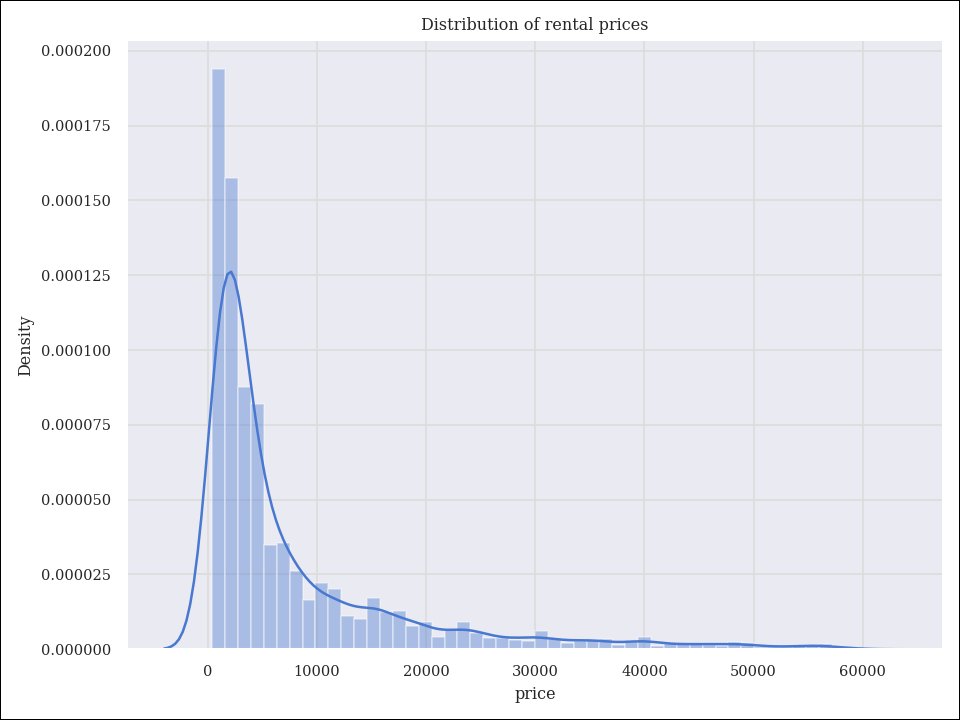}
    \caption{Distribution of rental prices in Ghana}
    \label{fig:fig1}
\end{figure}
Figure \ref{fig:fig1} reveals a right-skewed distribution of rental prices in Ghana during the last quarter of 2024, indicating a concentration of properties in lower price ranges with a few high-priced outliers. 

\begin{figure}[H]
    \centering
    \includegraphics[width=1\linewidth]{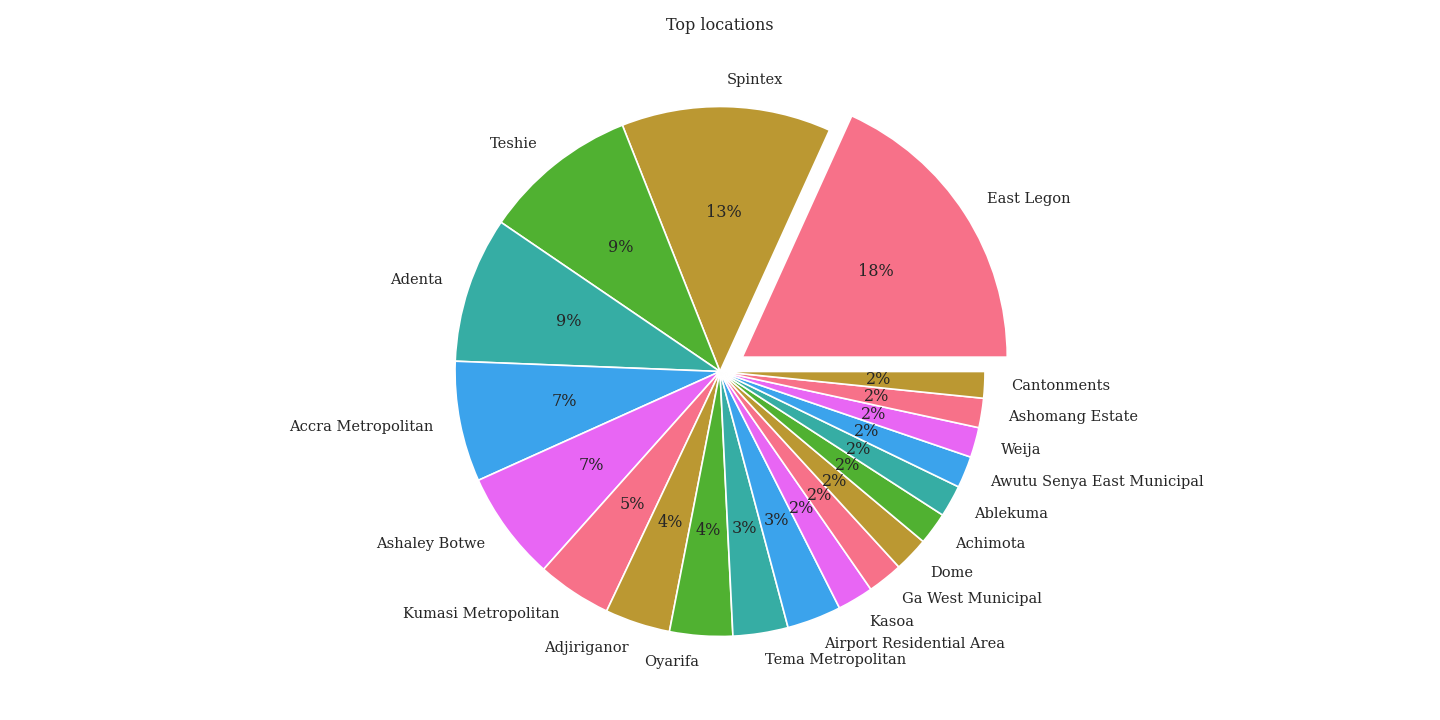}
    \caption{Distribution of rental listings in different locations in Ghana}
    \label{fig:fig2}
\end{figure}

Figure \ref{fig:fig2}  shows the geographical distribution of rental listings, with East Legon (18), Spintex (13), and Teshie/Adenta (9) having the highest concentrations. The remaining locations hold smaller proportions (2-7\%).

\begin{figure}[H]
    \centering
    \includegraphics[width=1\linewidth]{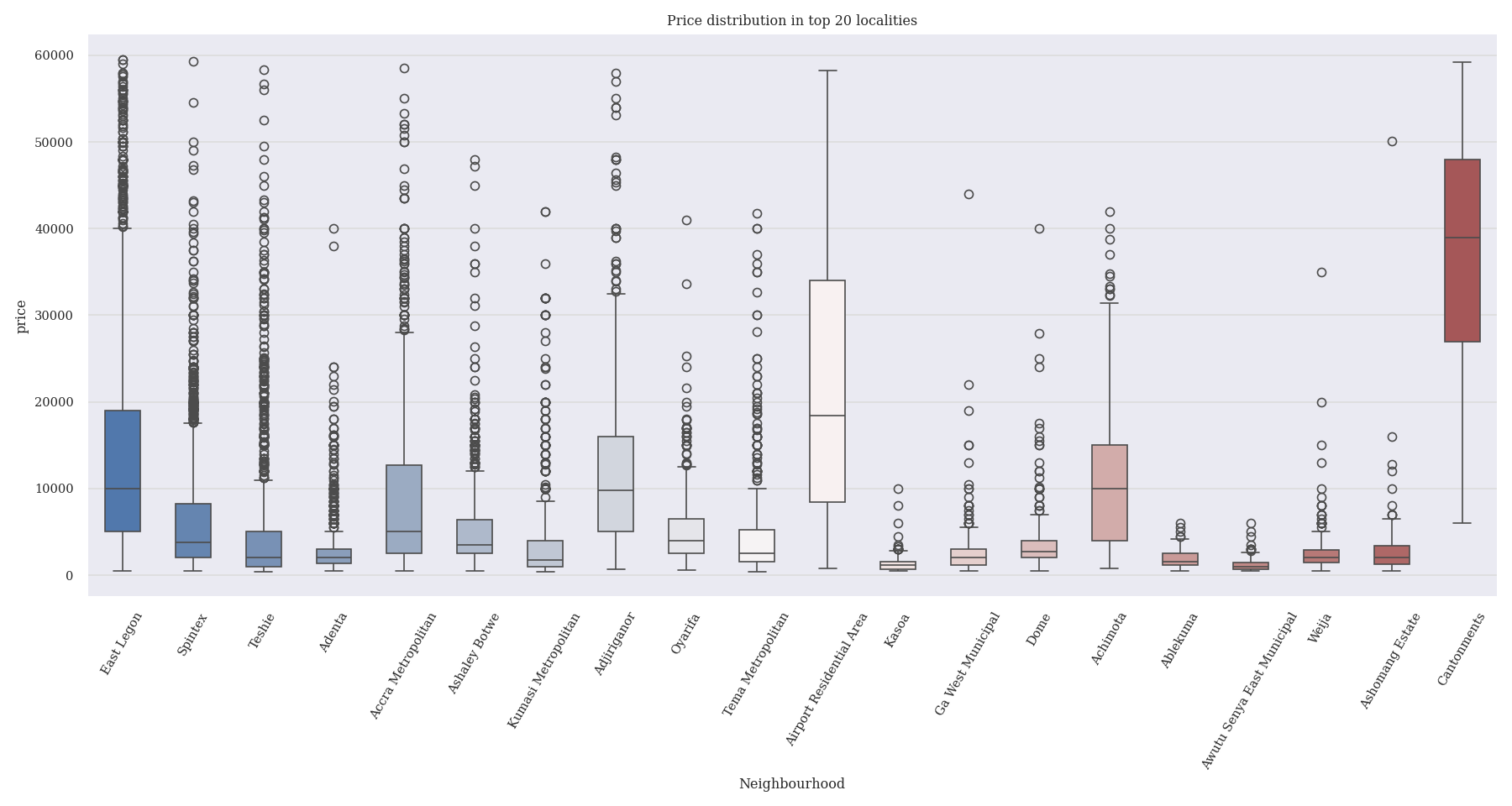}
    \caption{Illustration of rental prices in different locations in Ghana}
    \label{fig:fig3}
\end{figure}

The box plot in Figure \ref{fig:fig3} illustrates the variability in rental prices across different locations. Cantonments, East Legon, and Spintex have the highest median prices, with notable outliers indicating some very expensive properties in these areas. 

\begin{figure}[H]
    \centering
    \includegraphics[width=1\linewidth]{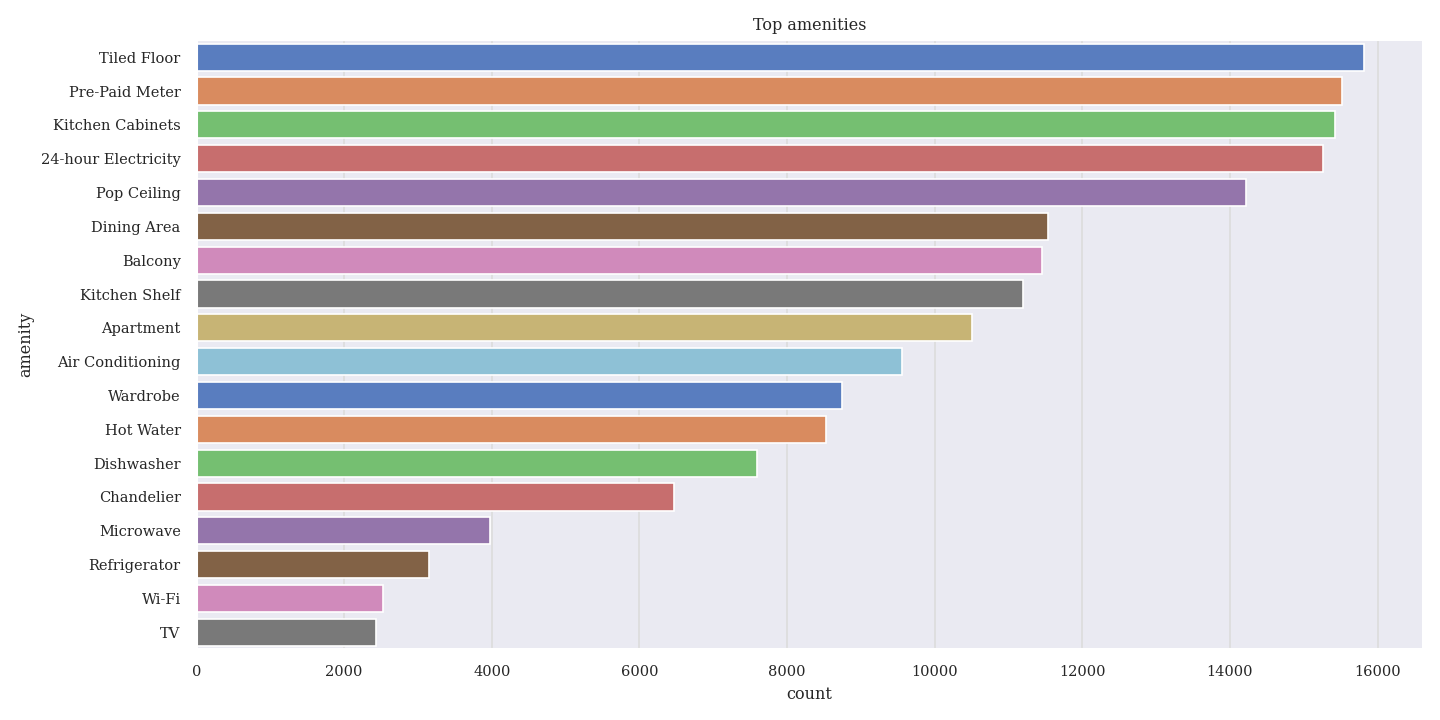}
    \caption{Distribution of top amenities provided in rental houses}
    \label{fig:fig4}
\end{figure}

Finally, Figure \ref{fig:fig4} presents the distribution of amenities, showing that tiled floors, prepaid meters, and kitchen cabinets are common, while Wi-Fi and TVs are less prevalent, potentially indicating their premium status in the rental market.

\section{Results}

From Table \ref{tab:table1}, it shows the performance of various regression techniques in predicting rental prices. CatBoost outperformed all other models, achieving the highest \(R^2\) score ($0.877$) and the lowest MSE ($0.162$), RMSE ($0.403$), and MAE ($0.301$). Linear Regression had the least performance, while SVR, Random Forest, and XGBoost performed competitively, with improvements in metrics compared to Linear Regression.

\begin{table}[H]
    \centering
    \renewcommand{\arraystretch}{1.5} %
    \begin{tabular}{|c|c|c|c|c|c|} \hline 
         \bfseries Regression technique & \bfseries $R^2$ &  \bfseries MSE &  \bfseries RMSE &  \bfseries MAE & \bfseries CV Score \\ \hline 
         Linear Regression&  0.749 &  0.331 &  0.575 &  0.436 & 0.745 \\ \hline 
         SVR&  0.819 &  0.239 &  0.489 &  0.361&  0.787 \\ \hline 
         Random Forest&  0.859 &  0.186 &  0.431 &  0.318 & 0.824 \\ \hline
 XGBoost& 0.868 & 0.174 & 0.418 & 0.312 &0.836 \\ \hline 
         \textbf{CatBoost}&  \textbf{0.877} &  \textbf{0.162} &  \textbf{0.403} &  \textbf{0.301} & \textbf{0.854} \\ \hline
    \end{tabular}
    \caption{Comparative performance of baseline models}
    \label{tab:table1}
\end{table}
The results in Table \ref{tab:table1} demonstrate that advanced ensemble methods like CatBoost and XGBoost consistently outperform traditional techniques such as Linear Regression and SVR in predicting rental prices. CatBoost achieved the best performance across all metrics, highlighting its ability to handle categorical data and capture complex relationships in the dataset effectively. Furthermore, its high cross-validation score suggests strong generalisation ability, indicating the model is likely to perform well on new, unseen data. This makes CatBoost a promising choice for predicting rental prices in this context.

To further enhance model performance, hyperparameter optimisation was performed. The results of this optimisation are presented in the following table, highlighting the impact of fine-tuning on predictive accuracy.

\begin{table}[H]
    \centering 
    \renewcommand{\arraystretch}{1.5} %
    \begin{tabular}{|c|>{\raggedright\arraybackslash}p{5cm}|c|c|c|c|c|} \hline
         \bfseries Regression technique &  \bfseries Best Parameters &  \bfseries $R^2$ &  \bfseries MSE &  \bfseries RMSE & \bfseries MAE & \bfseries CV Score \\ \hline 
         Linear Regression &  \textit{copy\_x=True, fit\_intercept=False, positive=False} &  0.749 &  0.331 &  0.575 &  0.436 & 0.745 \\ \hline 
         SVR &  \textit{kernel='rbf', gamma=0.1, C=1} &  0.822&  0.234 &  0.484 & 0.356 & 0.786 \\ \hline 
         Random Forest &  \textit{n\_estimators=333, max\_depth=90, max\_features='sqrt'} &  0.858 &  0.187 &  0.433 &  0.322 & 0.826 \\ \hline 
         XGBoost &  \textit{gamma=0.5, max\_depth=2, min\_child\_weight=2, sub\_sample=0.4} &  0.852 &  0.195 &  0.442 &  0.335 & 0.838 \\ \hline 
         \textbf{CatBoost} &  \textit{\textbf{learning\_rate=0.03, l2\_leaf\_reg=1, iterations: 1000, depth=8}} &  \textbf{0.876} &  \textbf{0.164} &  \textbf{0.404} &  \textbf{0.301} & \textbf{0.854} \\ \hline
    \end{tabular}
    \caption{Impact of hyperparameter optimisation on model performance.}
    \label{tab:table2}
\end{table}
Table \ref{tab:table2} summarises the performance of various regression models after hyperparameter tuning, highlighting their optimised parameters and evaluation metrics. CatBoost maintained its position as the best-performing model, achieving the highest \(R^2\) score of 0.876 and the lowest MSE, RMSE, and MAE values, demonstrating its ability to make highly accurate predictions. Hyperparameter tuning did not significantly alter the overall ranking of the models. CatBoost continues to be the top-performing model with the highest accuracy and robustness. While some models (like SVR) showed minor improvements, others (like XGBoost and Random Forest) experienced slight performance degradation.
\subsection{Model Validation }

To validate CatBoost's performance, we analysed its residual plots. Residuals represent the difference between actual and predicted rental prices. Ideally, residuals should be randomly distributed around zero, indicating no systematic bias in the model's predictions. We examined residual plots after applying an inverse log transformation to the target variable (rental price). This transformation helps address potential skewness in the price distribution and allows for a more accurate assessment of model performance. By comparing the distribution of residuals, we can gain further insights into the model's behaviour and identify any potential issues. We will further analyse the prediction error, which quantifies the difference between predicted and actual values, to assess the model's accuracy and identify any patterns in the errors.
\begin{figure}[H]
    \centering
    \includegraphics[width=0.75\linewidth]{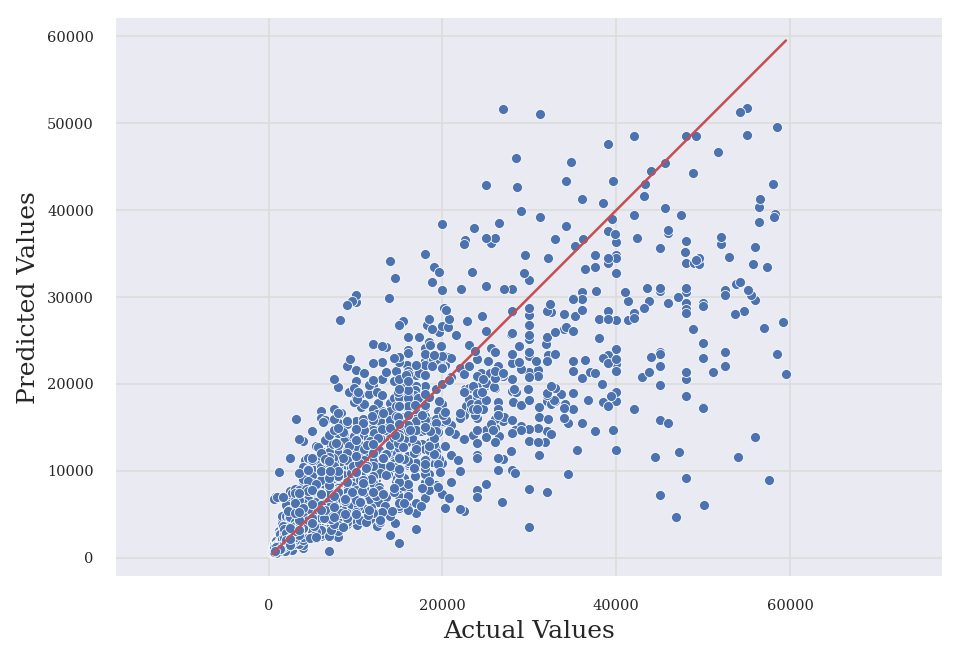}
    \caption{Scatter plot of actual vs predicted rental prices}
    \label{fig:fig5}
\end{figure}
The regression plot in Figure \ref{fig:fig5} shows a strong positive correlation between the actual rental prices and the predicted rental prices. The majority of the data points cluster around the diagonal line, indicating that the model's predictions are generally accurate. However, there are a few outliers where the predicted values deviate significantly from the actual values.

This plot visually confirms the model's ability to capture the underlying relationship between the features and the target variable. The presence of outliers suggests that there might be some instances where the model struggles to accurately predict the rental price, potentially due to complex interactions between factors or the presence of rare or unusual data points.
\begin{figure}[H]
    \centering
    \includegraphics[width=1\linewidth]{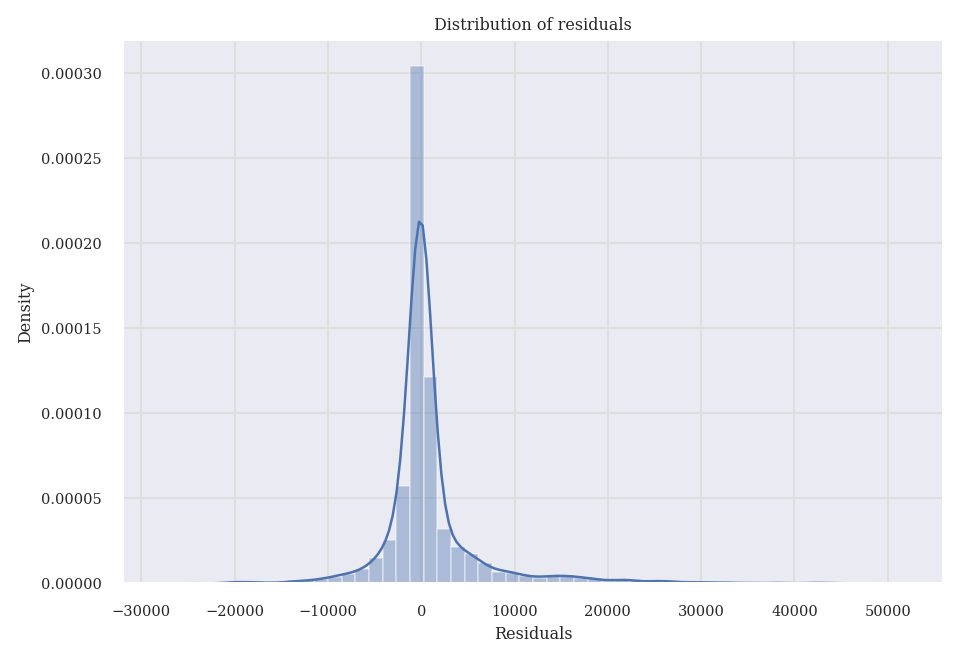}
    \caption{Histogram of actual predicted rental prices.}
    \label{fig:fig6}
\end{figure}
The residual plot in Figure \ref{fig:fig6} reveals the distribution of errors (residuals) between the predicted and actual rental prices. The residuals are centred around zero, with a relatively symmetrical bell-shaped distribution. This indicates that the CatBoost model generally provides unbiased predictions, as there is no systematic overestimation or underestimation across the dataset. The symmetry of the residuals is a positive sign, suggesting that the assumptions of linearity and normality in the regression model are largely satisfied.

However, the presence of longer tails on both sides of the distribution indicates that there are some instances where the model produces larger errors. These could be due to extreme outliers in the data or specific variations in rental prices that are not well-represented by the features used in the model. Such cases may suggest the need for further refinement, such as better feature engineering, more robust handling of outliers, or exploring additional external factors that influence rental prices. While the residual distribution supports the overall validity of the model, addressing these larger errors could enhance its predictive accuracy.
\begin{figure}[H]
    \centering
    \includegraphics[width=1\linewidth]{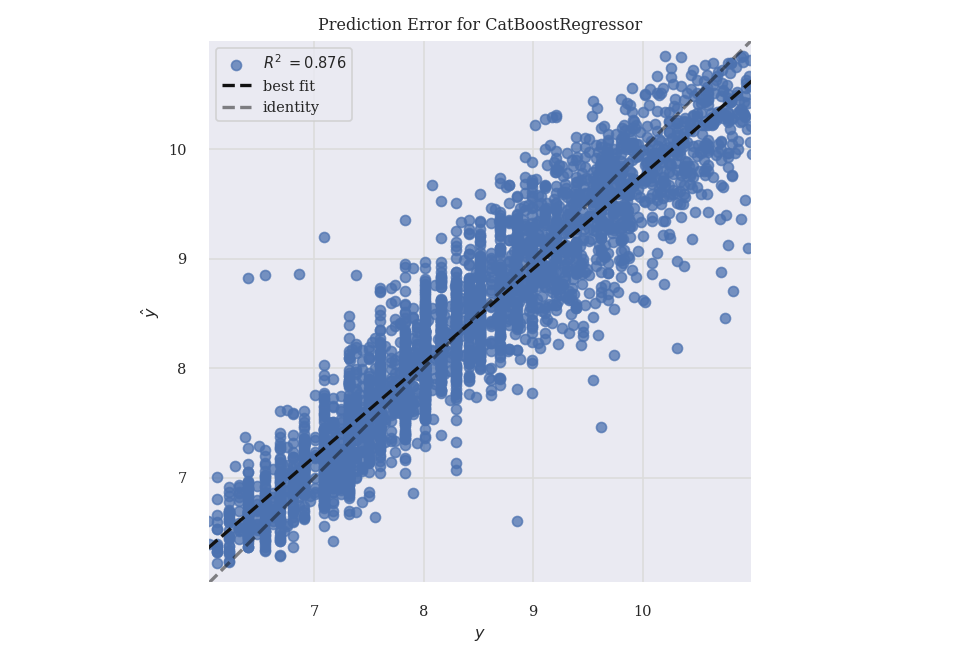}
    \caption{Scatter plot of prediction errors}
    \label{fig:fig7}
\end{figure}

The prediction error plot in Figure \ref{fig:fig7} for the CatBoost model shows the relationship between the actual values ($y$) and the predicted values ($\hat{y}$). The points are densely clustered around the identity line (dashed line), which indicates a strong alignment between the predicted and actual values. The \(R^2\) value of $0.876$ further demonstrates that the model explains a high proportion of the variance in the target variable, suggesting that the CatBoost model performs well in predicting rental prices.

However, some scatter can be observed, especially further from the centre, indicating that prediction errors increase slightly for extreme values. This could be attributed to the presence of outliers or under-represented patterns in the data. Despite these deviations, the best-fit line (solid line) closely tracks the identity line, reinforcing the effectiveness of the CatBoost model in capturing the underlying relationships in the data. Additional optimisation or feature engineering could further reduce prediction errors for outliers.

To summarise, the residual analysis and prediction error plots confirm the overall validity and effectiveness of the CatBoost model in predicting house rental prices in Ghana. 
\subsection{Feature Importance}

To understand how CatBoost makes predictions, we analysed feature importance. This analysis reveals which features have the most significant impact on rental price predictions. Features like location, number of bedrooms, and amenities are expected to be highly influential. Explaining feature importance enhances model interpretability and provides valuable insights into the key drivers of rental prices in the Ghanaian housing market. Visualising feature importance, such as through a bar chart, can effectively communicate these insights. 
\begin{figure}[H]
    \centering
    \includegraphics[width=1\linewidth]{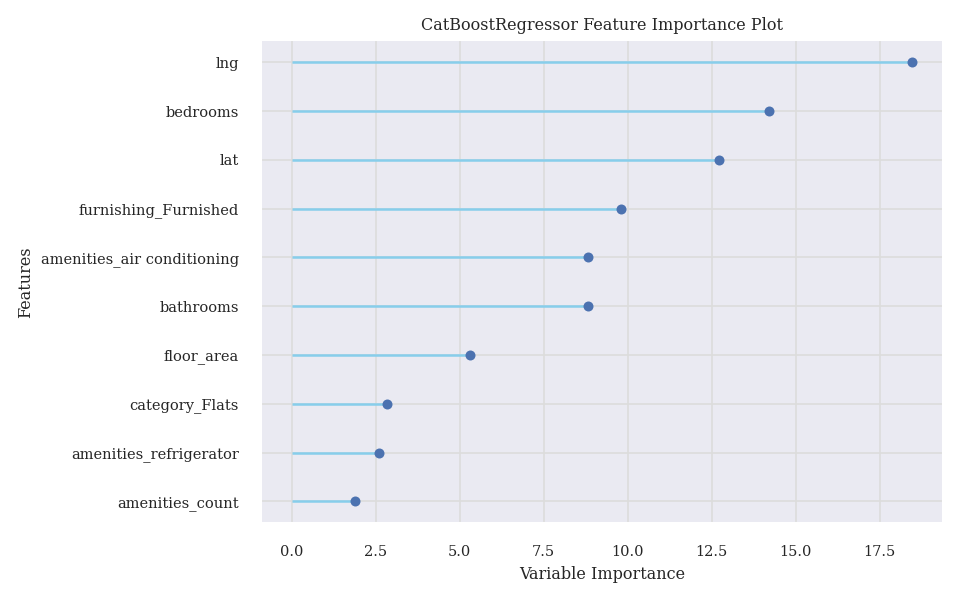}
    \caption{Feature importance with CatBoost}
    \label{fig:fig8}
\end{figure}

The feature importance plot in Figure \ref{fig:fig8} provides insights into the factors that most strongly influence rental prices in the model. Notably, location features ("lng" and "lat"), dominate the plot, indicating that geographical location is a primary driver of rental prices. This is expected, as factors like proximity to amenities, transportation hubs, and desirable neighbourhoods significantly impact property values. The number of bedrooms and bathrooms also rank high in importance, aligning with established real estate principles. The presence of the furnishing feature suggests that furnished properties are considered a valuable feature. Specific amenities like Air conditioning and refrigerator also contribute to price variations. On the other hand, house types has a relatively lower importance, suggesting that the property type (flats) might have a less significant impact on price compared to other factors.

It's important to remember that feature importance scores represent the individual importance of each feature. However, interactions between features might also influence price. For example, the relationship between "floor area" and "location" could be significant. Additionally, the specific importance of features might vary depending on the local market conditions and the target audience.
\section{Conclusion}

This study provides a comprehensive analysis of house rental price prediction in Ghana using machine learning, with CatBoost emerging as the most effective model, achieving an \(R^2\) of $0.876$. These findings suggest that ensemble-based machine learning models are well-suited for addressing real-world housing price prediction challenges, particularly in dynamic markets like Ghana.

Our analysis reveals key drivers of rental prices, notably location-based features, number of bedrooms, bathrooms, and furnishing status. While specific amenities contribute, their influence is less pronounced than location and property characteristics. The visualisations, including residual analysis and feature importance plots, offer valuable insights for stakeholders such as real estate professionals, investors, and policymakers. 
\subsection{Limitations and Further Study}

This study acknowledges certain limitations that provide avenues for future research. The model's performance is contingent on the available data, which may not fully represent Ghana's diverse housing market. Expanding the dataset with more features, such as property age, proximity to amenities, and neighbourhood characteristics, could enhance predictive accuracy. Additionally, incorporating temporal data to capture price fluctuations over time would be beneficial. 

Further research could explore advanced modelling techniques, including deep learning architectures, to potentially improve performance. Investigating regional variations within Ghana could also provide valuable insights. Finally, addressing potential biases in the data and model is crucial for ensuring fairness and generalisability.

\subsection{Additional Information}
\begin{itemize}
\item \textbf{Source Code}:  \href{https://github.com/epigos/house-prices-prediction}{https://github.com/epigos/house-prices-prediction}
\item \textbf{Dataset}: \href{https://www.kaggle.com/datasets/epigos/ghana-house-rental-dataset/}{https://www.kaggle.com/datasets/epigos/ghana-house-rental-dataset/}
\item \textbf{Funding}: This research received no external funding.
\item \textbf{Informed Consent Statement}: Not applicable.
\item \textbf{Conflict of Interest}: The author(s) declare no conflicts of interest related to this analysis or the use of the tools and techniques discussed.
\end{itemize}

\printbibliography
\end{document}